\documentclass[letter]{article}

\usepackage{fullpage}
\usepackage[english]{babel}
\usepackage[utf8]{inputenc}
\usepackage[T1]{fontenc}
\usepackage{amsmath,amsthm,amssymb,amsfonts,mathrsfs,latexsym,stmaryrd,mathtools}
\usepackage{indentfirst}
\usepackage{caption}
\usepackage{subcaption}
\usepackage{color,xcolor,graphicx,textpos,array,comment,wrapfig,hyperref}

\title{SPARQL Generation with Entity Pre-trained GPT for KG Question Answering}

\author{%
	Diego Bustamante$^{1,2}$,
	Hideaki Takeda$^{3}$\\ \\
	Authors are listed in alphabetical order.\\	
	{$^1$ Department of Computer Science, PUC, Santiago, Chile}\\
	{$^2$ Millenium Institute for Foundational Research on Data, Santiago, Chile}\\
	{$^3$ National Institute of Informatics (NII), Tokyo, Japan}
}

\begin{document}
	\setcounter{page}{1}
	
	\maketitle

	\begin{abstract}
		Knowledge Graphs popularity has been rapidly growing in last years. All that knowledge is available for people to query it through the many online databases on the internet. Though, it would be a great achievement if non-programmer users could access whatever information they want to know. There has been a lot of effort oriented to solve this task using natural language processing tools and creativity encouragement by way of many challenges. Our approach focuses on assuming a correct entity linking on the natural language questions and training a GPT model to create SPARQL queries from them. We managed to isolate which property of the task can be the most difficult to solve at few or zero-shot and we proposed pre-training on all entities (under CWA) to improve the performance. We obtained a 62.703\% accuracy of exact SPARQL matches on testing at 3-shots, a F1 of 0.809 on the entity linking challenge and a F1 of 0.009 on the question answering challenge.
	\end{abstract}

	\section{Introduction}
	
	Knowledge Graphs (KG) \cite{kg-book} are structures often used to represent knowledge. Nowadays, more than ever, the amount of open source information stored in KG throughout the web is beyond comprehension for us humans, e.g., just Wikidata has more than 108 million items\footnote{Wikidata:Statistics - \url{https://www.wikidata.org/wiki/Wikidata:Statistics}}.
	
	A clever way to store Knowledge Graphs are graph databases \cite{kg-book}. This approach solves a lot of important problems regarding information management, some of them are: storage, structure, query and update. Although access points to this data are readily available, for example via web browser, information extraction is not a simple task for a user that does not know how to make graph databases queries given a particular language (e.g. SPARQL\footnote{SPARQL 1.1 Query Language - \url{https://www.w3.org/TR/sparql11-query/}}).
	
	With the recent rise of generative LLMs it is natural for a user to think that ChatGPT\footnote{OpenAI ChatGPT - \url{https://chat.openai.com/}}, Copilot\footnote{Microsoft Copilot - \url{https://copilot.microsoft.com/}}, Bard\footnote{Google Bard - \url{https://bard.google.com/}} or any other model, could help them in the task of creating a SPARQL query in order to answer a question about a particular KG. Taking in consideration that generating queries it is not the main interest of the mentioned LLMs, the problem is that a simple qualitative analysis using this AI tools can expose that only a few of them do well on this task.
	
	The goal of this work is to identify the main property that makes this problem a difficult one, and propose a training methodology and model architecture such to improve the accuracy of this type of models on this particular task.

	\section{Related Work}
	
	There has been a lot of effort from researchers on improving results on the task of Knowledge Graph Question Answering. Given a KG, the problem of KGQA consists of inputs such as ``What are the papers written by the person X?'' and outputs such as the triples of the KG that correspond to the answer to the question. With the goal of motivating competition and innovation, research groups have published challenges and datasets with varied difficulties. Some of the most recent data collections on the field of scholarly knowledge are DBLP-QuAD \cite{awale_2023_7643971} and SciQA \cite{auer_2023_7744048}.
	
	The Scholarly QALD challenge recently took place at the ISWC 2023\footnote{Scholarly QALD at ISWC 2023 - \url{https://kgqa.github.io/scholarly-QALD-challenge/2023/}}. Many groups had great results and, as expected, most of them used LLMs in some aspect of their approaches \cite{NLQxform, bertology, leveragingLLMs, subgraph}.
	
	Since Natural Language Processing tools have demonstrated being very useful solutions for text-to-text tasks \cite{DBLP:journals/corr/abs-1810-04805, DBLP:journals/corr/VaswaniSPUJGKP17}, it is natural to think where can those tools be integrated to tackle the inputs of KGQA. The solutions seems to be to considerate SPARQL queries as plain text, as a natural language with its own vocabulary and grammar. Then, technologies like GPT \cite{DBLP:journals/corr/VaswaniSPUJGKP17} are a clear study path to solve the Scholarly QALD challenge.

	We manage to replicate the results obtained by Rajpal \& Usbeck \cite{bertology}. We studied how to improve the lack of multi-hops after entity linking in their approach. Inspired by the LLM popularity, we manage to isolate the most difficult characteristic of this task, create a training process to solve it and propose future improvements to our approach.	

	\section{Approach}	

	Our approach uses the DBLP-QuAD dataset \cite{awale_2023_7643971}. This data is composed of 10,000 items, every data point has a question, paraphrased question, SPARQL query, entities, relations and answer (triples). This structure enable us to approach the challenge in many different ways, stimulating innovation.
	
	Instead of directly trying to obtain the triples to answer the question, we first focused on dividing the task and translating the natural language question to a SPARQL query.
	
	\subsection{Entity linking}
	
	Our first models did not do very well. Our hypothesis was that translating a sub-string entity into a link while trying to learn the SPARQL grammar was too much work for our model size (3.47 M parameters) and for the amount of training data we had (10,000 items).
	
	To solve that problem, we subdivided the task into entity linking and SPARQL generation. We assumed the entity linking task can be done with perfect accuracy and then we would perform the translation. We build a new dataset where the entities are replaced by their iri, e.g., if the question was ``What are the papers written by the person Wei Li?'', the new question will be ``What are the papers written by the person <https://dblp.org/pid/64/6025-131>?''. This reduced the amount of different tokens (vocabulary) from more than 46,000 to only 10,399; a significant amount that can affect how the model will perform. We used the entity linking methods of Rajpal \& Usbeck \cite{bertology, banerjee2023dblplink} with less overhead (and worse accuracy).
	
	We ruled out the TP61 template \cite{awale_2023_7643971} data points due to error detected in the SPARQL queries, entities were not well used on the template. Since the final challenge can certainly have TP61-type questions, this decision is a source of error. But, our goal is to find new training ways to improve performance and comparatively will negatively affect all our models the same. 
	
	From 20,000 items (questions and paraphrased questions), we ended up with 9,289 successful entity linked data points on the new dataset. We minimize type I error, i.e., we did many checks to ensure perfect entity linking. In production or real testing we would bet on doing it right to improve accuracy, but in this phase we are building a new training dataset. Is evident that our EL performance is not state-of-the-art, but it was not our goal.
	
	The difference between the original and the entity linked dataset can are more clear in \autoref{dataset_table}. We also performed normalization on the questions described in \autoref{sec:experiments}.
	
	\begin{table}
		\centering
		\small
		\begin{tabular}{||c>{\raggedright}p{5.2cm}>{\raggedright}p{7.6cm}||} 
			\hline
			Dataset & Input example & Output example\tabularnewline [0.5ex] 
			\hline\hline
			Original & Show the Wikidata ID of the person Robert Schober. & SELECT DISTINCT ?answer WHERE \{ <https://dblp.org/pid/95/2265> <https://dblp.org/rdf/schema\#wikidata> ?answer \}\tabularnewline
			\hline
			Entity linked & show the Wikidata ID of the person <https://dblp.org/pid/95/2265> & SELECT DISTINCT ?answer WHERE \{ <https://dblp.org/pid/95/2265> <https://dblp.org/rdf/schema\#wikidata> ?answer \}\tabularnewline
			\hline
			Pre-train & <https://dblp.org/pid/95/2265> <https://dblp.org/pid/95/2265> ... & <https://dblp.org/pid/95/2265> <https://dblp.org/pid/95/2265> ...\tabularnewline
			\hline
		\end{tabular}
		\caption{Input and output examples for the GPT model by dataset. The original dataset (20,000 items), the entity linked dataset (9,289 items) and the pre-train dataset (7,617 items).}
		\label{dataset_table}
	\end{table}
	
	\subsection{Query generation}
	
\begin{center}
		\begin{figure}[!htbp]
			\begin{subfigure}{0.50\columnwidth}
				\centering
				\includegraphics[scale=0.085]{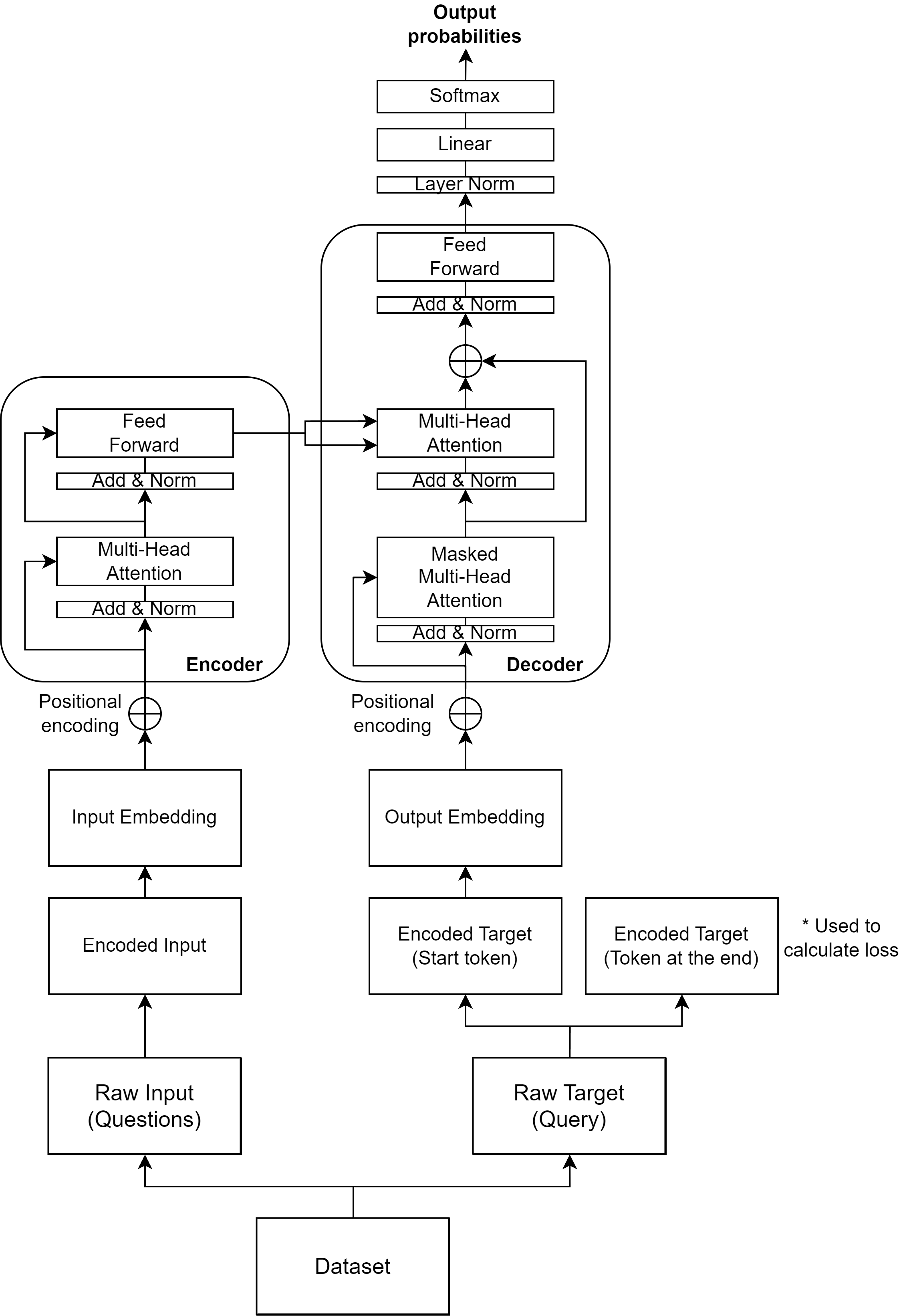}
				\caption{Diagram of our GPT architecture, very similar to the original model.}
				\label{fig:transformer}
			\end{subfigure}
			\hfill
			\begin{subfigure}{0.45\columnwidth}
				\centering
				\includegraphics[scale=0.07]{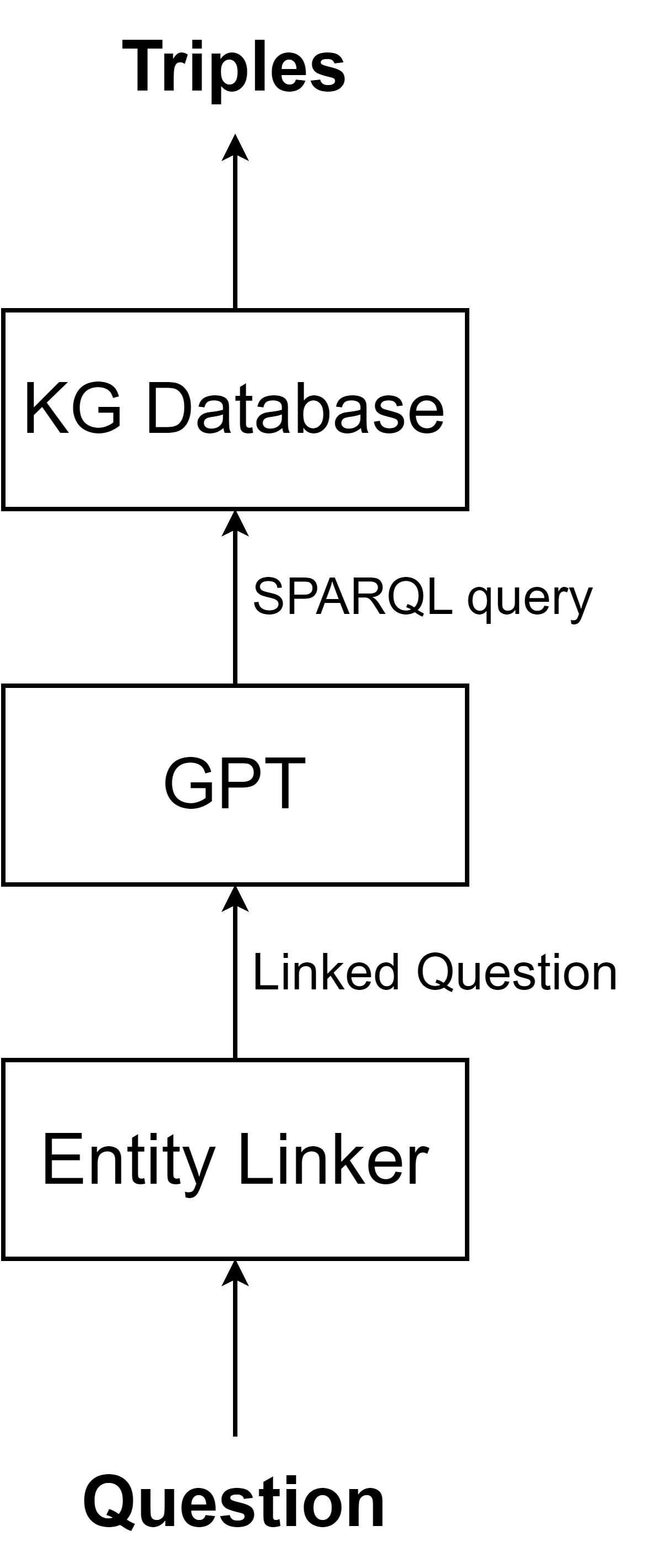}
				\caption{Complete process when every phase is integrated, we focused on GPT.}
				\label{fig:process}
			\end{subfigure}
			\caption{Our solution, the components of our model and how to integrate it in production setting.}
			\label{fig:approach}
		\end{figure}
	\end{center}

	With the entity link phase done, a translation model has to learn the different types of query templates and complete the variable information with data from the input. The architecture of our model is based on the GPT implementation by Andrej Karpathy\footnote{Let's build GPT: from scratch, in code, spelled out - \url{https://youtu.be/kCc8FmEb1nY}} which is, in turn, inspired on the paper ``Attention Is All You Need'' \cite{DBLP:journals/corr/VaswaniSPUJGKP17}. Karpathy implemented a decoder only transformer so we completed the encoder-decoder model to take the questions as inputs on the encoder and the decoder would generate the queries. The complete architecture is shown in \autoref{fig:transformer}.
	
	As we said, the model has to learn $i)$ query templates and $ii)$ learn how to change the template with the corresponding information. Simplifying the second task, it can be described as an identity function $f(string) = string$ with the entity iris found on the input. The problem with our first solution was that while learning the grammatical aspects of the problem it stopped from learning the rest. Our GPT preferred a lazy minimization of the amount of wrong tokens than learning how to perform the entity replacement. We attribute this behavior to the fact that there are only a few entities per question/query so its contribution to the loss function is very small. The parameter optimizer fitted a shortcut learning \cite{DBLP:journals/corr/abs-2004-07780} and could not improve more.
	
	It is well studied that pre-training can improve the performance of generative transformer models \cite{alyafeai2020survey, raffel2023exploring, DBLP:journals/corr/abs-1810-04805, Radford2018ImprovingLU}, specially when post fine-tuning for question answer is implemented \cite{Radford2018ImprovingLU}. Therefore, based on literature \cite{alyafeai2020survey, raffel2023exploring}, our hypothesis is that performance should improve when we first train the model to perform the identity function on entities and then transfer that learning over the KGQA task.
	
	The amount of different queries is exponentially large, hence the need for an AI model to do the translation. In contrast, under the Closed World Assumption \cite{kg-book} the amount of entities is fixed. Then, we pre-train the model with all entities to perform the identity task. In the first instance we tried to teach the model the function $f(entity) = entity$, but since entities can be anywhere in the input and output the improvement was not meaningful. When we pre-train the function $f(entity\ entity\ entity\ ...) = entity\ entity\ entity\ ...$, we saw an important performance enhancement.
	
	Our GPT implementation can be found at GitHub\footnote{Code - \url{https://github.com/DiegoEmilio01/SPARQL-generation-with-pre-trained-GPT-for-KG-Question-Answering/}}. When combined with an entity linker and the desired KG database to query, the complete process looks like \autoref{fig:process}.
	
	\section{Experiments \label{sec:experiments}}
	
	In addition to the entity linking, we also modify the questions to achieve a better generalization in the model. We eliminate punctuation symbols and first word capitalization, and spaced parenthesis because our tokenization is by words space-split. For example, ``<https://dblp.org/pid/64/6025-131>?'' should be the same entity as ``<https://dblp.org/pid/64/6025-131>''.
	
	The distribution of the 9,289 data points is 8,648 for training (93.1\%), 456 for validation (5.9\%) and 185 for testing (2\%). This proportions were found by trial and error, it is important to notice that we did not follow the original distribution due to the creation of the new entity linked dataset. In the case of the pre-training, all 7,617 entities were used as train, validation and testing. The original dataset had 7,143 entities but we added new iris from the final 500 questions of the Scholarly QALD challenge to the vocabulary justified by our Closed World Assumption \cite{kg-book}. The average number of entities per query is 1.231 entities.
	
	Example of items in the all datasets are tabulated in \autoref{dataset_table}. In the case of the pre-train dataset, the length of the inputs are 34 tokens and outputs are 49 tokens, this is, the maximum input and output size of our model.
	
	Our vocabulary was originally of size 9,339; but with the inclusion of the final 500 questions it ended being of size 10,399 (including new entities and other words). The model hyperparameters were also found by trial and error. Dropout of 1\%, learning rate of 0.0007, internal vector of dimension 128, number of heads of 8, and 4 layers for the decoder and encoder each. We instanced 2 models, one with pre-training and one without to compare improvements. Both were of size 3.47 million parameters, were trained the same total amount of epochs (19,200) and with the exact same data points distribution between training, validation and testing, in order to be able to compare them. In the case of the pre-trained model, it was trained 14,400 epochs on entities and 4,800 on the new dataset.
	
	As training metric we used cross entropy for the loss function and as testing metric we used accuracy at 1 shot, accuracy at 3 shots and average Hamming distance. Accuracy in our setting means how many exact SPAQRL queries could we generate over the total, and Hamming distance is how many tokens away is our prediction from the gold answer (only one shot). For the final 500 questions of the challenge we retrieved final triples as answers and submitted them to the official challenge\footnote{[DBLP-QuAD] Scholarly QALD @ ISWC 2023 - \url{https://codalab.lisn.upsaclay.fr/competitions/14264}} to obtain precision, recall and F1 (under the user DiegoEmilio).
	
	\section{Results and Discussion}
	
	\begin{table}
		\centering
		\begin{tabular}{||c c c c c c c||} 
			\hline
			Model $\backslash$ Metric & Acc@1(\%) & Acc@3(\%) & aHD(\# tokens) & Precision(\%) & Recall(\%) & F1\\ [0.5ex] 
			\hline\hline
			Not pre-trained & 31.892 & 43.784 & 1.724 & 0.502 & 0.605 & 0.005 \\
			\hline
			Pre-trained & 49.189 & 62.703 & 2.011 & 0.725 & 1.291 & 0.009 \\
			\hline
		\end{tabular}
		\caption{Results for the 2 types of models. The metrics are query accuracy at 1 shot, query accuracy at 3 shots, query average Hamming distance, challenge precision, challenge recall and challenge F1.}
		\label{experiment_table}
	\end{table}

	Our results are shown in \autoref{experiment_table}. Pre-training the model improves almost every metric. A 17.3\% improvement on accuracy@1 is remarkable, tough precision, recall and F1 are not much significant because getting few questions with a lot of triples right can produce a random increase in F1 like ours. We observed that the model without pre-training tried to minimize how many tokens gets wrong, i.d., improve the average Hamming distance. In contrast, with pre-training it learned how to maximize the number of exact match answers despite getting more tokens wrong when generating an incorrect answers (hence the higher aHD).
	
	We have overcome the initial shortcut learning \cite{DBLP:journals/corr/abs-2004-07780}. This suggests that changes in the loss function could help the training. For example, reward higher exact matches because in our experiments the reward difference between getting 1 to 2 tokens incorrect is the same as between 0 to 1.
	
	Due to the type of model and training, we expected our model to extrapolate the rules to manage queries it has never seen before \cite{extrapolate}. Considering our accuracy in testing, we think that it managed to learn the different templates. But, the identity function is still a problem when the author has never been seen in training.
	
	The architecture of our model allows it to be very good at few-shot \cite{DBLP:journals/corr/abs-2005-14165}, and we attribute to this property the success of our results with such little data points (9,289). We tried to generate zero-shots queries with entities we knew were not in the dataset. Even when all iris were included in the pre-training, the model was not capable of generate the exact match, just the template. Moreover, when including the final 500 questions to the vocabulary we added 474 new entities to the problem, so this hints that the reason behind our bad results on the challenge is zero-shots.
	
	This analysis is also backed up by our metrics. The 4th column on \autoref{experiment_table} shows that, on average, the model gets tokens wrong approximately the same amount of tokens as the number of entities per query (1.231). Thus, considering that entities are the context in which the question should be answered, we identify that Comprehension-topic hallucinations \cite{hallucinations} of the identity function at zero-shots is the main difficulty of this task.
	
	Our approach has the potential to be refined and combined to improve results, for example is heavily dependent on the performance of the entity linking phase before the GPT phase. Our entity linking process achieved 92.931\% precision, 71.635\% recall and 0.80905 F1. Best entity linkers could be used to improve performance but since we are competitive with state-of-art this should not bring a high increase in performance.
	
	We propose a better training to tackle zero-shots inaccuracies. Since our framework is under the Closed World Assumption \cite{kg-book}, we can add few queries per entity on top of our pre-training. This enrichment of the dataset should be enough to improve performance, specially on the final 500 questions of the Scholarly QALD challenge.
	
	Furthermore, we think that the value of our results lies in at least returning the correct template for difficult questions and queries. In a qualitative analysis, most of the popular LLMs can't help users to formulate queries as hard as the ones in the DBLP-QuAD dataset \cite{awale_2023_7643971}. In that sense, being competitive with most popular LLMs using a very small model and training data in comparison shows an optimistic path to solve this task.
	
	Finally, a light model like ours can be trained with very few data, time and costs. When a new open source Knowledge Graph appears on the internet is not known by LLMs. Then, the publishers of the KG can find this method useful to easily help their non-programmer users by putting in production a GPT model that suggests queries.
	
	\section*{Acknowledgement}
	
	The research was partially funded by the National Institute of Informatics, Japan. D.B. acknowledges partial support from ANID, Subdirección de Capital Humano (Magíster Nacional, 2023, folio 22231282).
	
	Big thanks to Trinidad Gatica and Elías Sabja for their help with code debugging. 
	
	\bibliographystyle{acm}
	\bibliography{report.bib}

\end{document}